\documentclass[10pt,twocolumn,letterpaper]{article}

\usepackage{iccv}
\usepackage{times}
\usepackage{epsfig}
\usepackage{graphicx}
\usepackage{amsmath}
\usepackage{amssymb}

\usepackage[utf8]{inputenc} 
\usepackage[T1]{fontenc}    
\usepackage{url}            
\usepackage{booktabs}       
\usepackage{amsfonts}       
\usepackage{nicefrac}       
\usepackage{microtype}      
\usepackage{multirow}
\usepackage{multicol}
\usepackage[ruled]{algorithm2e}
\usepackage{verbatim}
\usepackage{bm}
\usepackage[switch]{lineno}
\usepackage[table,xcdraw]{xcolor}

\usepackage{enumitem}
\usepackage[accsupp]{axessibility}  

\newcommand{\B}{\mathcal{B}}
\newcommand{\C}{\mathcal{C}}
\newcommand{\D}{\mathcal{D}}
\newcommand{\F}{\mathcal{F}}
\newcommand{\I}{\mathcal{I}}

\renewcommand{\P}{\mathcal{P}}
\newcommand{\X}{\mathcal{X}}
\newcommand{\R}{\mathcal{R}}

\newcommand{\Z}{\mathcal{Z}}

\newcommand{\qileft}{[\kern-0.15em[}
\newcommand{\qiLeft}{\left[\kern-0.32em\left[}
\newcommand{\qiright}{]\kern-0.15em]}
\newcommand{\qiRight}{\right]\kern-0.32em\right]}

\newcommand{\qiLeftm}{\left[\kern-0.28em\left[}
\newcommand{\qiRightm}{\right]\kern-0.28em\right]}

\newcommand{\qiLefts}{\left[\kern-0.16em\left[}
\newcommand{\qiRights}{\right]\kern-0.16em\right]}

\usepackage[pagebackref=true,breaklinks=true,letterpaper=true,colorlinks,bookmarks=false]{hyperref}

\iccvfinalcopy 

\def\iccvPaperID{10686} 

\ificcvfinal\fi

\begin{document}

\title{Re-mine, Learn and Reason: Exploring the Cross-modal Semantic Correlations for Language-guided HOI detection}

\author{Yichao Cao\\
Southeast University\\
{\tt\small caoyichao@seu.edu.cn}
\and
Qingfei Tang\\
Nanjing Enbo Tech.\\
{\tt\small qingfeitang@gmail.com}
\and
Feng Yang\\
Southeast University\\
{\tt\small yangfeng@seu.edu.cn}
\and
Xiu Su\thanks{Corresponding author.}\\
University of Sydney\\
{\tt\small xisu5992@uni.sydney.edu.au}
\and
Shan You\\
SenseTime\\
{\tt\small youshan@sensetime.com}
\and
Xiaobo Lu\\
Southeast University\\
{\tt\small xblu@seu.edu.cn}
\and
Chang Xu\\
University of Sydney\\
{\tt\small c.xu@sydney.edu.au}
}

\maketitle
\ificcvfinal\thispagestyle{empty}\fi

\begin{abstract}

Human-Object Interaction (HOI) detection is a challenging computer vision task that requires visual models to address the complex interactive relationship between humans and objects and predict <$human, action, object$> triplets. Despite the challenges posed by the numerous interaction combinations, they also offer opportunities for multi-modal learning of visual texts. In this paper, we present a systematic and unified framework (\textbf{RmLR}) that enhances HOI detection by incorporating structured text knowledge. Firstly, we qualitatively and quantitatively analyze the loss of interaction information in the two-stage HOI detector and propose a re-mining strategy to generate more comprehensive visual representation. Secondly, we design more fine-grained sentence- and word-level alignment and knowledge transfer strategies to effectively address the many-to-many matching problem between multiple interactions and multiple texts. These strategies alleviate the matching confusion problem that arises when multiple interactions occur simultaneously, thereby improving the effectiveness of the alignment process. Finally, HOI reasoning by visual features augmented with textual knowledge substantially improves the understanding of interactions. Experimental results illustrate the effectiveness of our approach, where state-of-the-art performance is achieved on public benchmarks.

\end{abstract}

\section{Introduction}
\label{sec:intro}

Human-object interaction (HOI) detection \cite{gupta2015visual, chao2018learning} is an emerging field of research that builds upon object detection and requires more advanced high-level visual understanding. A high-performing HOI detector should not only accurately localize all interacting Human-Object pairs but also recognize their specific interactions, typically represented as an HOI triplet in the format of <$human, action, object$> \cite{zhou2021cascaded}.

Previous approaches for achieving HOI detection can be divided into two pipelines: those that treat object detection and interaction recognition as separate stages \cite{zhang2022efficient, chao2018learning, gao2020drg, gao2018ican, li2019transferable, hou2021detecting}, and those that aim to handle both simultaneously \cite{gkioxari2018detecting, kim2020uniondet, zhong2021glance, liao2020ppdm, chen2021reformulating}. Although both paradigms have made significant progress, the task remains challenging due to the vast variety of human-object interaction combinations in the real world \cite{yuan2022rlip, yuan2022detecting}. For example, the HICO-DET dataset \cite{chao2018learning} contains 600 human-object interaction combinations. A common approach is to optimize the model by mapping these various triplet labels into a discrete one-hot labels. However, this method oversimplifies the intricacy of the HOI task and can be cumbersome for model optimization.

In recent years, multi-modal learning has gained significant attention in the vision-and-language learning domain, where it has achieved state-of-the-art performance on various tasks \cite{jing2023deep,baltruvsaitis2018multimodal, cao2022searching, li2022grounded, alayrac2022flamingo,jing2021amalgamating}. By integrating information from multiple modalities, such as images \cite{su2021bcnet,su2021prioritized,su2022vitas,su2021locally} and text \cite{zheng2023can}, multi-modal learning can provide a more comprehensive understanding of entities or events. In the field of HOI, several recent studies \cite{zhong2021polysemy, iftekhar2022look, li2022improving, wang2022learning, yuan2022rlip, yuan2022detecting} have applied image-and-text models to improve interaction detection performance. For example, HOI-VP \cite{zhong2021polysemy} used a set of binary classifiers to verify each category and proposed Language Prior-guided Channel Attention (LPCA) to enhance HOI recognition. SSRT \cite{iftekhar2022look} pre-selected object-action (OA) prediction candidates and encoded them as text features to refine the queries' representation. PhraseHOI \cite{li2022improving} employed a pre-trained word embedding model to generate a phrase embedding that enhances the discriminative ability and capacity of the common knowledge space.

Although the use of vision-and-language pre-training (VLP) or language knowledge injection has motivated the exploration of HOI image-text correspondences through multi-modal learning, their effectiveness in knowledge transfer remains limited. This is due to the heterogeneity gap \cite{baltruvsaitis2018multimodal} that exists between different modalities, which requires cross-modal modeling to reduce the inter-modality gap and explore semantic correlations. Additionally, the problem of multi-interaction to multi-text matching in HOI tasks remains unsolved, which may limit the reliability of cross-modal correspondences. Therefore, a systematic and unified solution is needed to better exploit cross-modal HOI detection and improve the generalization ability of HOI detectors.

In this paper, we propose a systematic approach (RmLR) to improve HOI detection in light of the structured text knowledge in cross-modal learning. Concretely, our HOI framework proceeds from three perspectives: \emph{i)} we reveal the problem of interaction information loss in the two-stage HOI detector, and propose the \textbf{\emph{Re-mine}} strategy to obtain this crucial visual information; \emph{ii)} more sophisticated cross-modal \textbf{\emph{Learning}} method to achieve semantic association from sentence- and word-level; \emph{iii)} \textbf{\emph{Reasoning}} using textual knowledge-enhanced representations substantially improves the visual model's understanding of interactions. The main contribution of this paper is summarized as follows:
\begin{itemize}
\item We propose a systematic and unified framework so that the inherent challenges of HOI can be elaborated in both visual and cross-modal settings.

\item We qualitatively and quantitatively analyze the problem of interaction information loss in two-stage visual HOI detector, and propose a re-mining strategy to capture these crucial interaction-aware representations.

\item We formulate the cross-modal learning in HOI domain as a many-to-many matching problem, where multiple interactions need to be matched with their corresponding textual descriptions, and propose appropriate sentence and text alignment strategies to promote learning semantically aligned.

\item  Extensive experiments show that our RmLR equipped with ResNet-50 outperforms previous SOTA by a large margin and achieves an average mAP increase of about +3.88p and +5.05p on HICO-DET \cite{chao2018learning} and V-COCO \cite{gupta2015visual}, respectively.

\end{itemize}

\section{Related Work}

\subsection{Generic HOI Detection}
According to the network architecture design, current HOI detection approaches can be broadly classified into two categories: two-stage methods \cite{zhang2022efficient, chao2018learning, gao2020drg, gao2018ican, li2019transferable, hou2021detecting} and one-stage methods \cite{gkioxari2018detecting, kim2020uniondet, zhong2021glance, liao2020ppdm, chen2021reformulating}. One-stage methods typically employ multitask learning to jointly perform instance detection and interactive relation modeling \cite{liao2020ppdm, zhang2021mining, liao2022gen}. In contrast, two-stage methods first perform object detection, followed by interactive relation modeling for all HO pairs candidates. By leveraging the full potential of each module, two-stage methods have demonstrated improved detection performance \cite{zhang2021mining}. Recent works have also leveraged the power of Transformer \cite{carion2020end} in formulating HOI detection as set prediction, resulting in significant performance gains \cite{chen2021reformulating}.

\begin{figure}[t]
\centering
\includegraphics[width=0.99\columnwidth]{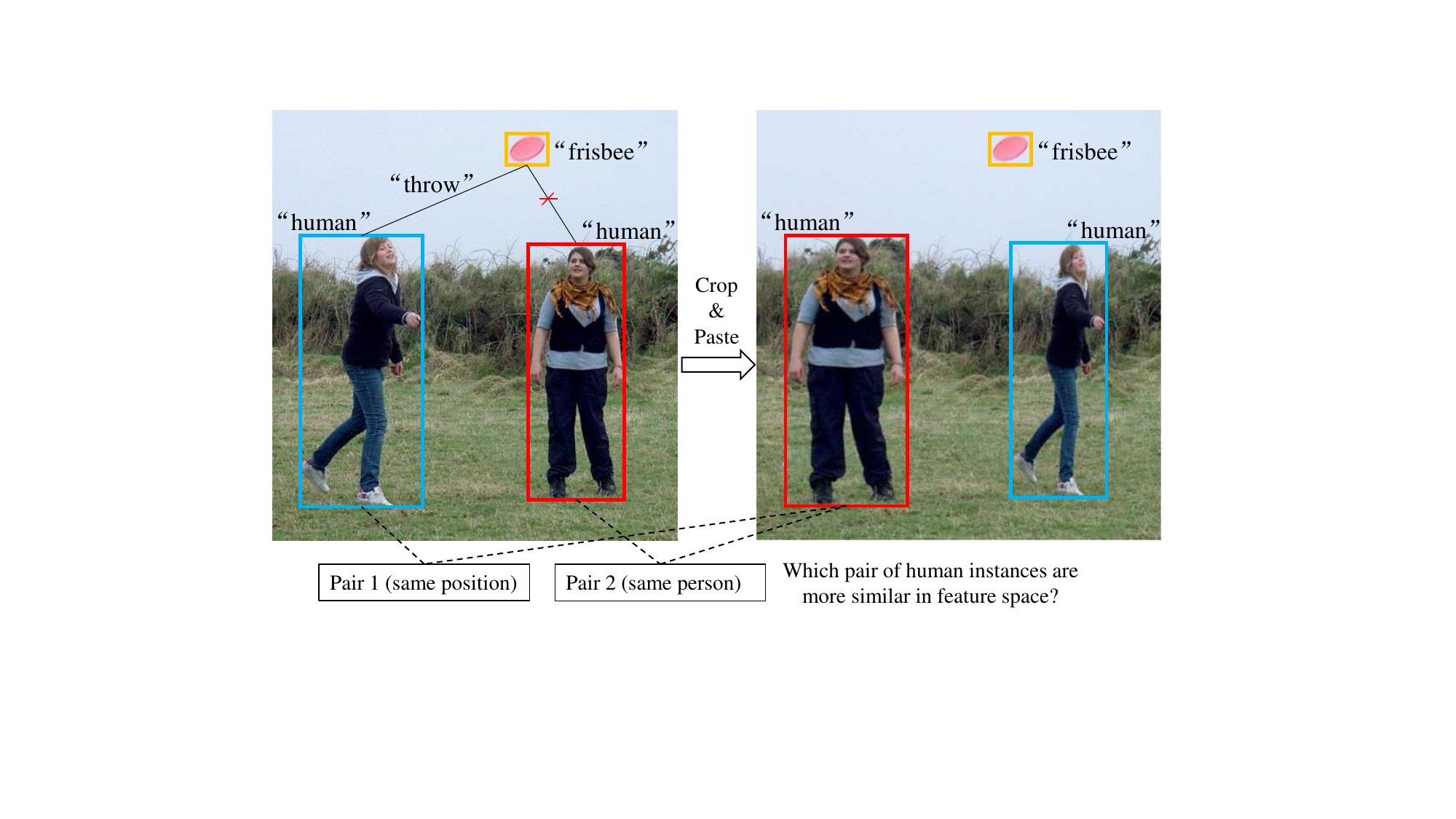} 
\caption{Which pair of human instances is more similar in the HOI detector? According to our analysis using cosine similarity measurement for the human tokens of the DETR-based HOI detector \cite{zhang2022efficient}, Pair 1 has a similarity score of 0.99, while Pair 2 has a score of only 0.58. These findings are consistent with numerous similar cases observed in our experiments, highlighting the phenomenon of \emph{interaction-related information loss} in which the output tokens of the object detector primarily emphasize spatial position, potentially leading to the loss of crucial information related to the interactions.}
\label{fig1}
\end{figure}

\subsection{Language Semantics for Vision}
Motivated by the remarkable success of Large Language Model (LLM) \cite{devlin2018bert} pre-training in NLP, leveraging language semantics to enhance vision models has recently emerged as a promising approach for computer vision tasks \cite{radford2021learning, tsimpoukelli2021multimodal, jing2022learning, alayrac2022flamingo, zhang2022glipv2, jing2021meta}. Among them, Vision-and-Language Pre-training (VLP) \cite{mu2021slip, li2022grounded} has become a popular paradigm in many vision-and-language tasks due to its applicability in learning generalizable multi-modal representations from large-scale image-text data \cite{alayrac2022flamingo, chen2022align, Cheng_2022_CVPR}. These methods have been recently used in multi-modal retrieval \cite{dzabraev2021mdmmt}, vision-and-language navigation \cite{anderson2018vision}, and other fields. Effective inter-modal semantic alignment, especially fine-grained semantic alignment, is a critical component for cross-modal learning \cite{li2022grounded}. Since different modalities have their own inherent properties, their semantic organization varies to some extent \cite{chen2021new}. Thus, it is crucial to investigate how to efficiently correlate diverse semantic information.

\begin{figure*}[t]
  \centering
  \includegraphics[width=0.99\linewidth]{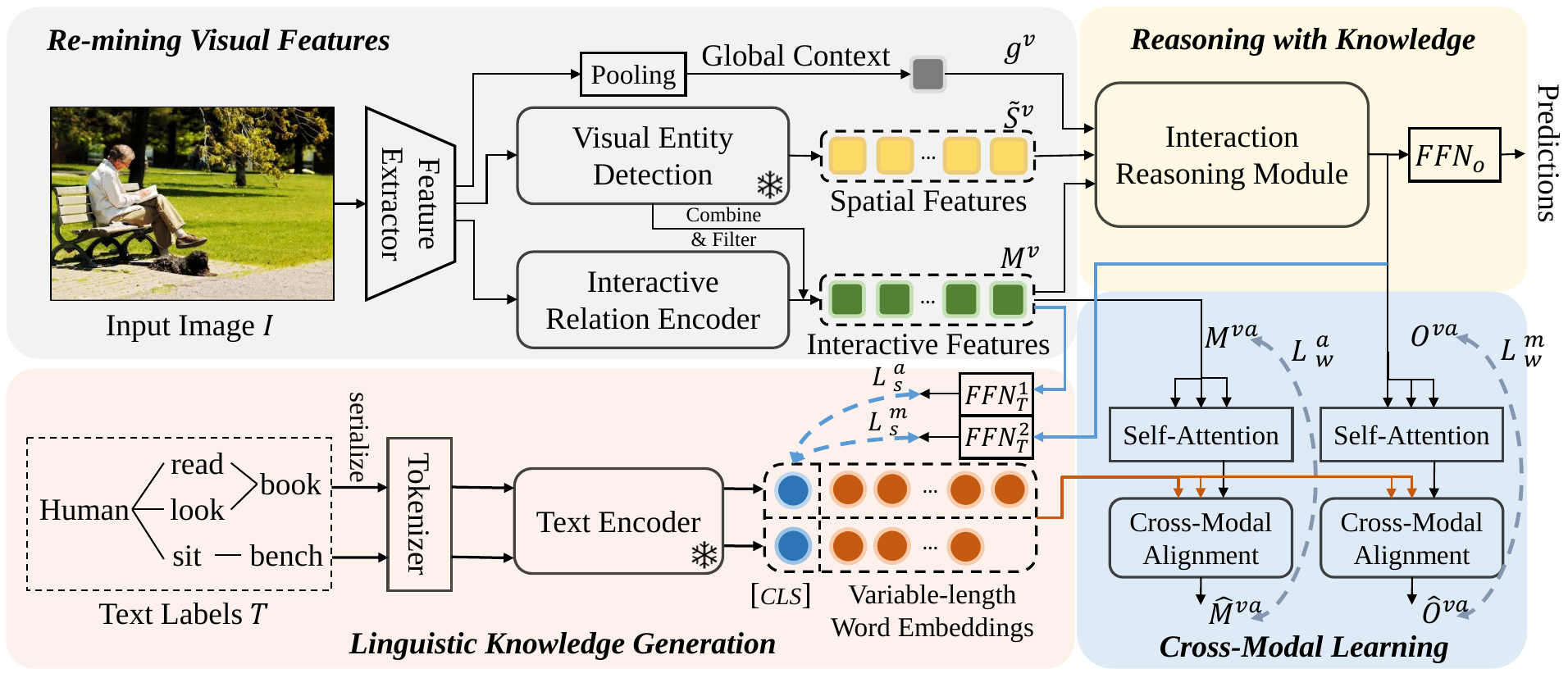}
   \caption{The overall architecture of our proposed RmLR approach, where the Visual Entity Detection module, Interactive Relation Encoder (with the “re-mining visual feature” process), Linguistic Knowledge Generation, Cross-Modal Learning (with the “learning cross-modal content” process), Interaction Reasoning Module (with the “reasoning using knowledge” process) are shown.}
   \label{fig:2}
\end{figure*}

\subsection{HOI Vision-and-Language Modeling (HOI-VLM)}
\label{sec:sec2.3}
Although previous HOI detectors \cite{zhang2021mining, zhang2022efficient, liu2022interactiveness}  have achieved moderate success, they often treat interactions as discrete labels and ignore the richer semantic text information in triplet labels. More recently, a few researchers  \cite{zhong2021polysemy, iftekhar2022look, li2022improving, wang2022learning, yuan2022rlip, yuan2022detecting} have investigated the HOI Vision-and-Language Modeling to further boost the HOI detection performance. Among them, \cite{zhong2021polysemy}, \cite{iftekhar2022look}, and \cite{yuan2022detecting} both tended to aggregate language prior features into the HOI recognition. RLIP \cite{yuan2022rlip} and \cite{wang2022learning} proposed to construct a transferable HOI detector via the VLP approach. As the applications and extensions of Vision-and-Language learning to the HOI domain, these HOI-VLM methods aim to understand the content and relations between visual interaction features and their corresponding triplet texts. However, the natural distribution inconsistency in the two modalities can directly lead to incompatibility of the modal features, as discussed by \cite{chen2021new}. The issue of narrowing the heterogeneity gap and effectively ensuring the consistency and correlation of cross-modal features in HOI detection remains unresolved.


\section{The Proposed RmLR Framework}
\subsection{Overview Architecture}


We adopt the two-stage HOI detector approach for its superior performance, interpretability, and intuitive intermediate features. Inspired by the DETR family \cite{carion2020end}, we design the RmLR architecture (see Figure \ref{fig:2}). Formally, our RmLR model is trained on an image-text corpus $\X=\left\{\left(\I^i, \mathcal{T}^i\right)\right\}_{i=1}^{|\mathcal{X}|}$, where $\I$ denotes the input image and $\mathcal{T}$ represents all the phrase descriptions (\emph{e.g.} ``Human ride bicycle'') in $\I$. We can roughly divide RmLR into visual feature learning module $\Phi_{\theta_\mathcal{V}}$, interaction reasoning module $\Phi_{\theta_\R}$ and a pre-trained text encoder $\Phi_{\theta_\mathcal{T}}$, where $\theta$ indicates the weights in different modules. The overall training objective is defined as follows,
\begin{equation}
\min \mathbb{E}_{(\I, \mathcal{T}) \sim \X} \left[ \mathcal{L} (\mathcal{GT},\Phi_{\theta_\mathcal{T}}(\mathcal{T}), \Phi_{\theta_\mathcal{V}} \circ \Phi_{\theta_\mathcal{R}}(\I,\mathcal{Q}_o))\right]
\end{equation}
where $\mathcal{GT}$ and $\mathcal{L}$ are ground-truth label and overall loss function respectively, $\mathcal{Q}_o$ denotes the set of queries of objects, and $\circ$ is a network compound operator. Details of the module implementation are explained in the subsequent sections.



\subsection{Re-mining Visual Features}
\label{sec:sec3.2}
\textbf{Visual Entity Detection} An input image $\I \in \R^{H \times W \times C}$ is first extracted as low-level visual features $\X^v \in \R^{h \times w \times c}$, and then the features are segmented into patch embeddings $\left\{x_1^v, x_2^v, \ldots, x_{N_v}^v\right\}$, where $N_v$ is the number of patch embeddings. Then the patch embeddings $\left\{x_1^v, x_2^v, \ldots, x_{N_v}^v\right\}$ are ﬂattened and linearly projected through a linear transformation $\mathcal{E}^v \in \R^{c \times D^v}$. Specifically, the input for Transformer-based entity detection are calculated via summing up the patch embeddings and position embeddings $\mathcal{E}_{p o s}^v \in \R^{N_v \times D^v}$:
\begin{equation}
\Z^v=\left[ x_1^v \mathcal{E}^v ; x_2^v \mathcal{E}^v ; \ldots ; x_{N_v}^v \mathcal{E}^v\right]+\mathcal{E}_{p o s}^v
\end{equation}
Through self-attention, cross-attention, and feed-forward network (FFN) inference in entity detection decoder $\F_{E D}$, we obtain the entity token features $\mathcal{S}^v \in \mathcal{R}^{N \times D^v}$, box locations $\B^v \in \R^{N \times 4}$ and instance classes $\C^v \in \R^{N \times N_c}$:
\begin{equation}
(\mathcal{S}^v, \B^v, \C^v)=\F_{E D}\left(\Z^v, \mathcal{Q}_o \right)
\end{equation}
where $N$ denotes the number of detected instances, $\mathcal{Q}_o$ denotes the set of queries of objects, and $N_c$ denotes the number of detectable categories. To obtain the pair-wise entity token features and box locations, we construct a set of pair-wise HO indexes $\{(h,o)\mid h \neq o, \C_h^v=\text {``human''}\}$. We form all pairs of detected instances and filter those where the subject is not human, as object–object pairs are beyond the scope of HOI detection. According to the filtered HO indexes, pair-wise entity token features $\tilde{\mathcal{S}}^v \in \R^{N^p \times 2D^v}$ and box locations $\tilde{\B}^v \in \R^{N^p \times 8}$ are able to obtain. This information is used for subsequent interactive relation learning and reasoning.

\begin{figure}[t]
  \centering
  \includegraphics[width=0.9\linewidth]{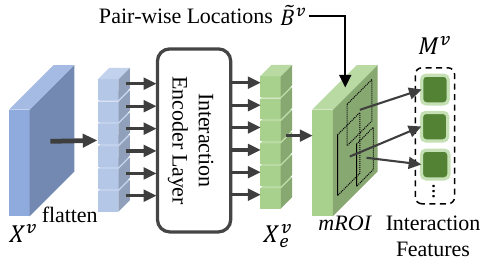}
   \caption{Re-mining the crucial interactive features via an interactive relation encoder.}
   \label{fig:IRE}
\end{figure}

\textbf{Interactive Relation Encoder} Through a meticulous analysis of numerous cases, we discovered that \emph{the current entity detection models prioritize the object's location information}. As a result, humans performing different actions at the same position are often mapped to similar representations, as illustrated in Figure \ref{fig1}. This phenomenon poses a significant risk to the HOI task, as it may result in the loss of crucial visual information. As two-stage HOI detectors operate independently for entity detection and interaction recognition, the entity token features $\mathcal{S}^v$ obtained from the entity detection model predominantly focus on spatial information and hence may fail to capture enough interaction-relevant cues.

\begin{figure}[t]
\centering
\includegraphics[width=0.9\linewidth]{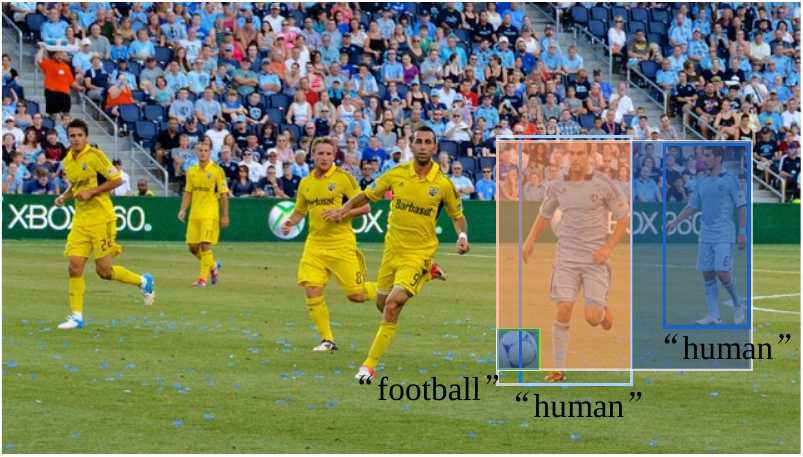} 
\caption{Feature interference problem in the naive union interaction region. According to the rules of the naive union interaction region, the {\color{orange}{orange}} and {\color{cyan}{blue}} part together constitute the interaction area of the rightmost person to the football. It can be seen that the interactive human-object features ({\color{orange}{orange}} part) interfere with the non-interactive human-object features ({\color{cyan}{blue}} part).}
\label{fig4}
\end{figure}

To this end, we design a lightweight Interactive Relation Encoder (IRE) to remine interaction features intuitively and explicitly (see Figure \ref{fig:IRE}). To capture the higher-level relation features from lower-level visual features, we apply a Transformer encoder to process feature map $\X^v$:
\begin{equation}
\X_e^v=\F_{enc}(\X^v)
\end{equation}
Then, we perform masked RoI operation on the interactive information-rich tensors $\X_e^v$ to compute the direct reflection $m^v$ according to the pair-wise box locations $\tilde{\B}^v$:
\begin{equation}
m^v={FC}({GAP}({mROI}(\X_e^v,\tilde{\B}^v))) \in \R^{D^v}
\end{equation}
Here, we use a fully-connected layer (${FC}$), global average pooling (${GAP}$), and masked region of interest (ROI) operation (${mROI}$) to obtain the interaction-aware features. To ensure that the features are only computed within the region of interest, we use a zero mask to cover the regions outside the HO candidate boxes to avoid the feature interference problem, as shown in Figure \ref{fig4}. After that, ${GAP}$ operation followed by an FC layer are applied on the feature map $\X^v$ to obtain global scene information $g^{v}$:
\begin{equation}
g^{v} = {FC}({GAP}(\X^v)) \in \R^{D^v}
\end{equation}
So far, we have generated the human and object candidates, global context $g^{v}$, pair-wise token $\tilde{\mathcal{S}}^v=\{\tilde{s}_i^v\}_{i=1}^{|\tilde{\mathcal{S}}^v|}$, and interaction cues ${\mathcal{M}}^v=\{{m}_j^v\}_{j=1}^{|{\mathcal{M}}^v|}$, which contain rich visual features for HOI recognition. The detailed ablations for this structure can be founded in Section \ref{sec:sec4.4} and Table \ref{tab1}.

\subsection{Linguistic Knowledge Generation}
To integrate linguistic knowledge into the visual HOI framework, we first construct annotation text for every image in HOI datasets. Considering the arrangement of <$person, verb, object$> triplet is very similar to the <$subject, predicate, object$> in language, we directly serialize each triplet annotation $\mathcal{GT}_i$ as a sub-sentence $t_i$. Then, a special $[SEP]$ token is used to separate multiple sub-sentences. In this way, each input image $\I$ obtains a corresponding variable-length annotation text $\mathcal{T}=\{t_j\}_{j=1}^{|\mathcal{T}|}$, where $|\mathcal{T}|$ denotes the number of ground truth interactions for the input image $\I$.

We utilize a pre-trained language model, such as MobileBERT \cite{sun2020mobilebert}, to generate semantic representations at the sentence- and word-level. First, the input text $\mathcal{T}$ is tokenized into subword tokens $\{x_1^l, x_2^l, \ldots, x_{N_l}^l\}$ using the WordPiece algorithm \cite{wu2016google}. These tokens are then represented as one-hot vectors $z_i^l \in \R^V$, where $V$ is the vocabulary size, and $N_l$ is the number of tokens. The tokens are then linearly transformed into embeddings using a matrix $\mathcal{E}^l \in \R^{V \times D^l}$. Additionally, a special start-of-sequence $[CLS]$ token embedding $z_{c l s}^l \in \R^{D^l}$ is added to the beginning of the text. Finally, the input text representations are obtained by summing up the token embeddings and text position embeddings $\mathcal{E}_{p o s}^l \in \R^{\left(N_l+1\right) \times D^l}$:
\begin{equation}
\X^l=\left[z_{c l s}^l  ; z_1^l \mathcal{E}^l ; \ldots ; z_{N_l}^l \mathcal{E}^l; z_{end}^l\right]+\mathcal{E}_{p o s}^l
\end{equation}
Using the text encoder $\F_{T E}$, we calculate the $[CLS]$ tokens $\mathcal{E}_{c l s}$ and word embeddings $\mathcal{E}^w$ as
\begin{equation}
\left(\mathcal{E}_{c l s}, \mathcal{E}^w\right)=\F_{T E}\left(\X^l\right)\in \R^{\left(N_l+1\right) \times D^l}
\end{equation}
In this way, the linguistic knowledge corresponding to $|\mathcal{T}|$  ground-truth interactions can be obtained, including sentence-level representation $\mathcal{E}_{c l s} = \{e_{cls}^i\}_{i=1}^{|\mathcal{T}|}$ and word-level representation $\mathcal{E}^w = \{e_{w}^j\}_{j=1}^{|\mathcal{E}^w|}$. We provide a detailed comparison of different text encoders in Section \ref{sec:sec4.4} and Table \ref{tab-text-encoder}.


\begin{table*}[]
\caption{Experimental results on HICO-DET [\textcolor{green}{6}] and V-COCO [\textcolor{green}{16}].}
\centering
\footnotesize
\begin{tabular}{cccccccc|cc}
\hline
            &            & \multicolumn{6}{c|}{HICO-DET}                                                        & \multicolumn{2}{c}{V-COCO} \\ \cline{3-10}
            &            & \multicolumn{3}{c}{Default   Setting} & \multicolumn{3}{c|}{Known   Objects Setting} &              &            \\ \cline{3-8}
Method (Year) & Backbone   & Full       & Rare      & Non-rare     & Full         & Rare         & Non-rare       &${AP}_{role}^{\# 1}$ & ${AP}_{role}^{\# 2}$   \\ \hline
One-stage Methods: \\
InteractNet (2018) \cite{gkioxari2018detecting}    &ResNet-50-FPN  &9.94   &7.16   &10.77  & -      & -& -&40.0& -\\
PPDM (2020) \cite{liao2020ppdm}                    &Hourglass-104  &21.94  &13.97  &24.32  &24.81  &17.09&27.12& -& -\\
HOTR (2021) \cite{kim2021hotr}                     &ResNet-50      &25.10  &17.34  &27.42  &       -&-&-&55.2&64.4\\
HOI-Trans (2021) \cite{zou2021end}                 &ResNet-101     &26.61  &19.15  &28.84  &29.13  &20.98&31.57&52.9&-\\
AS-Net (2021) \cite{chen2021reformulating}         &ResNet-50      &28.87  &24.25  &30.25  &31.74  &27.07&33.14&53.9&-\\
QPIC (2021) \cite{tamura2021qpic}                  &ResNet-101     &29.90  &23.92  &31.69  &32.38  &26.06&34.27&58.8&61.0\\
SSRT (2022) \cite{iftekhar2022look}                &ResNet-50      & 30.36  & 25.42  & 31.83   & -  & -  & - & 63.7 & 65.9 \\
SSRT (2022) \cite{iftekhar2022look}                &ResNet-101     & 31.34  & 24.31  & 33.32   & -  & -  & - & 65.0 & 67.1 \\
CDN-S (2022) \cite{zhang2021mining}                    &ResNet-50      & 31.44  & 27.39  & 32.64   & 34.09  & 29.63  & 35.42 & 61.68 & 63.77 \\
CDN-B (2022) \cite{zhang2021mining}                    &ResNet-50      & 31.78  & 27.55  & 33.05   & 34.53  & 29.73  & 35.96 & 62.29 & 64.42 \\
CDN-L (2022) \cite{zhang2021mining}                    &ResNet-101     & 32.07  & 27.19  & 33.53   & 34.79  & 29.48  & 36.38 & 63.91 & 65.89 \\
DOQ (CDN-S) (2022) \cite{qu2022distillation}       &ResNet-50      & 33.28  & 29.19  & 34.50   & -  & -  & - & - & - \\
Liu et al. (2022) \cite{liu2022interactiveness}    &ResNet-50      & 33.51  & 30.30  & 34.46   & 36.28  & 33.16  & 37.21 & 63.0 & 65.2 \\
GEN-VLKT-s (2022) \cite{liao2022gen}               &ResNet-50      & 33.75  & 29.25  & 35.10   & 36.78  & 32.75  & 37.99 &  62.41 & 64.46\\
GEN-VLKT-m (2022) \cite{liao2022gen}               &ResNet-101      & 34.78  & 31.50  & 35.77   & 38.07  & 34.94  & 39.01 & 63.28 & 65.58 \\
GEN-VLKT-l (2022) \cite{liao2022gen}               &ResNet-101      & 34.95  & 31.18  & 36.08   & 38.22  & 34.36  & 39.37 & 63.58 & 65.93 \\

\hline
Two-stage Methods: \\
HO-RCNN (2018) \cite{chao2018learning}             &CaffeNet       &7.81   &5.37   &8.54   &10.41  &8.94   &10.85  &-&-\\
GPNN (2018) \cite{qi2018learning}                  &ResNet-101    &13.11  &9.34   &14.23  & -      & -& -&44.0& -\\
TIN (2019) \cite{li2019transferable}               &ResNet-50     &17.03  &13.42  &18.11  &19.17  &15.51&20.26&47.8&54.2\\
VCL (2020) \cite{hou2020visual}                    &ResNet-50      &23.63  &17.21  &25.55  &25.98  &19.12&28.03&48.3&-\\
ATL (2021) \cite{hou2021affordance}                &ResNet-50      &23.81  &17.43  &27.42  &27.38  &22.09&28.96&-&-\\
VSGNet (2020) \cite{ulutan2020vsgnet}              &ResNet-152     &19.80  &16.05  &20.91  & -      & -& -&51.8&57.0\\
DJ-RN (2020) \cite{li2020detailed}                 &ResNet-50      &21.34  &18.53  &22.18  &23.69  &20.64&24.60& -& -\\
DRG (2020) \cite{gao2020drg}                       &ResNet-50-FPN  &24.53  &19.47  &26.04  &27.98  &23.11&29.43&51.0&-\\
IDN (2020) \cite{li2020hoi}                        &ResNet-50      &24.58  &20.33  &25.86  &27.89  &23.64&29.16&53.3&60.3\\
FCL (2021) \cite{hou2021detecting}                 &ResNet-50      &25.27  &20.57  &26.67  &27.71  &22.34&28.93&52.4&-\\
SCG (2021) \cite{zhang2021spatially}               &ResNet-50-FPN  &29.26  &24.61  &30.65  &32.87  &27.89&34.35&54.2&60.9\\
UPT (2022) \cite{zhang2022efficient}               &ResNet-50      & 31.66  & 25.90  & 33.36   & 35.05  & 29.27  & 36.77 & 59.0 & 64.5 \\
UPT (2022) \cite{zhang2022efficient}               & ResNet-101    & 32.31 & 28.55  & 33.44 & 35.65 & 31.60 & 36.86 & 60.7 & 66.2 \\

\rowcolor[HTML]{E0E0E0}
RmLR (Ours)                 & ResNet-50     & 36.93 & 29.03 & 39.29 & 38.29 & 31.41 & 40.34 & 63.78 & 69.81        \\
\rowcolor[HTML]{E0E0E0}
RmLR (Ours)                 & ResNet-101    & 37.41 & 28.81 & 39.97 & 38.69 & 31.27 & 40.91 & 64.17 & 70.23        \\
\hline
\end{tabular}
\label{tab2}
\end{table*}

\subsection{Cross-Modal Learning}


For visual representation, we first concatenate the global context $g^v$, pair-wise token $\tilde{s}^v = (s_h^v, s_o^v)$ and corresponding interaction cue $m^v$ to generate unified and diverse visual description for HO candidate:
\begin{equation}
\mathcal{H}^v={FC}(\operatorname{cat}\left(g^v,\tilde{s}^v, m^v\right)) \in \R^{D^l}
\end{equation}

Then, we introduce the competitive strategy in UPT \cite{zhang2022efficient} to construct a concise
Transformer-based interaction reasoning module $\F_{I R}$. After the competitive operation in $\F_{I R}$, the visual features $\mathcal{H}^v$ are converted into $\mathcal{O}^v$. To achieve a more flexible and efficient correlation of variable-length text to the interaction set, we design a dual distillation scheme to guide the training process for Interactive Relation Encoder and Interaction Reasoning Module simultaneously. Among them, the operation for IRE is more focused on pair-wise token $\tilde{s}^v$, and the latter is more focused on $\mathcal{O}^v$. The attention operation in these two mechanisms is defined as follows:
\begin{equation}
\mathcal{ATTN}(q, k, v)=\operatorname{softmax}\left(q k^{\top} / \sqrt{D_k}\right) \cdot v
\end{equation}
where $q$, $k$, and $v$ are the query, key, value matrices linearly transformed from the corresponding input sequences, respectively, and $D_k$ is the dimension of $k$. We conduct $L$ self-attention layers to interact representations within the two levels of features:
\begin{equation}
\mathcal{M}^{v s}=\mathcal{ATTN}\left(\mathcal{M}^v, \mathcal{M}^v, \mathcal{M}^v\right)
\end{equation}
\begin{equation}
\mathcal{O}^{v s}=\mathcal{ATTN}\left(\mathcal{O}^v, \mathcal{O}^v, \mathcal{O}^v\right)
\end{equation}
where $\mathcal{M}^{v s}$ and $\mathcal{O}^{v s}$ are the self-attention outputs for two representations, respectively. Then, the cross-modal attention are designed to align two modality representations and integrate linguistic information into visual representations in word-level:
\begin{equation}
\widehat{\mathcal{M}}^{v a}=\mathcal{ATTN}\left(\mathcal{M}^{v a}, \mathcal{E}^w, \mathcal{E}^w\right)
\end{equation}
\begin{equation}
\widehat{\mathcal{O}}^{v a}=\mathcal{ATTN}\left(\mathcal{O}^{v a}, \mathcal{E}^w, \mathcal{E}^w\right)
\end{equation}
where $\mathcal{M}^{v a}$ and $\mathcal{O}^{v a}$ are the visual representations corresponding to the textual embeddings $\mathcal{E}^w$, $\widehat{\mathcal{M}}^{v a}$ and $\widehat{\mathcal{O}}^{v a}$ are cross-attention outputs for two visual representations, respectively. In this way, no matter how complex multiple interaction are confronted, it is possible to align their visual features with the fine-grained textual representations. And the number of tokens of $\widehat{\mathcal{M}}^{v a}$ and $\widehat{\mathcal{O}}^{v a}$ are equal to the number of ${\mathcal{M}}^{v a}$ and ${\mathcal{O}}^{v a}$. In order to transfer linguistic knowledge to a visual model, we adopt the $L1$ distance metric to facilitate the learning between two types of representations:
\begin{equation}
\mathcal{D}_{L 1}\left({a}_{h o}, b_{h o}\right)=\frac{1}{N} \sum_i^N\left|a_{h o}-b_{h o}\right|
\end{equation}
where $a_{h o}$ and $b_{h o}$  broadly refer to two types of representations in our RmLR architecture. It is convenient to use word-level semantically enhanced representations to guide the learning of visual models. The two key components are guided as follows:
\begin{equation}
\begin{aligned}
\mathcal{L}_{w}^m = \mathbb{E}_{(\I, \mathcal{T}) \sim \X} \left[ \D_{L1}\left(\mathcal{O}^{va} , \widehat{\mathcal{O}}^{va}\right) \right]
\end{aligned}
\end{equation}

\begin{equation}
\mathcal{L}_{w}^a = \mathbb{E}_{(\I, \mathcal{T}) \sim \X} \left[ \D_{L1}\left(\mathcal{M}^{va} , \widehat{\mathcal{M}}^{va}\right)\right]
\end{equation}
where $\mathcal{L}_{w}^m$ and $\mathcal{L}_{w}^a$ denote the word-level cross-modal alignment loss for visual representation and logits, respectively. Even if multiple interactions occur between one HO pair, they can be described by variable-length word embedding sequences. These operations implement a fine-grained alignment and transfer mechanism for variable-length word embedding sequences to visual interaction set in HOI task.

In addition, we also perform sentence-level knowledge transfer for the RmLR. Although the sentence-level text representation is not as detailed as the word-level text representation, it also reflects the interaction information of HO pair to some extent. Thus, we regard sentence-level transfer as an auxiliary objective for our RmLR. Without the cross-modal attention, we directly perform knowledge transfer from [$CLS$] tokens $\mathcal{E}_{c l s}$ to the logits of RmLR:
\begin{equation}
\mathcal{L}_{s}^m = \mathbb{E}_{(\I, \mathcal{T}) \sim \X} \left[ \D_{L1}\left(\mathcal{E}_{c l s}, \mathcal{FFN}_T^2\left(\mathcal{O}^{va}\right)\right) \right]
\end{equation}
Similarly, we design an auxiliary objective on IRE, where the task is to guide the output representations of IRE:
\begin{equation}
\mathcal{L}_{s}^a = \mathbb{E}_{(\I, \mathcal{T}) \sim \X} \left[ \D_{L1}\left(\mathcal{E}_{c l s} , \mathcal{FFN}_T^1\left(\mathcal{M}^{va}\right)\right) \right]
\end{equation}

We also provided detailed ablation experiments and analysis of this structure in Table \ref{tab1} and Table \ref{tab-loss-weight} of Section \ref{sec:sec4.4}.

\subsection{Reasoning with Language-enhanced Representations}
\label{sec:sec3.5}
For the HOI recognition, a concise Transformer-based Interaction Reasoning Module (IRM) $\F_{I R}$ is designed to aggregate representation for each HO candidate. Our work differs from previous work in that these fed-in features are enhanced by textual knowledge, which is richer and more distinct than the unimodal features. After that, we add a classification head $\mathcal{FFN}_o$ to map logits to specific categories:
\begin{equation}
\P=\mathcal{FFN}_o(\F_{I R}(\mathcal{H}^v))
\end{equation}
Finally, a Focal loss is adopted as $\mathcal{L}_{\text {hoi }}$ to evaluate the image-level HOI predictions:
\begin{equation}
\mathcal{L}_{\text {hoi }} = Focal(sigmoid(\mathcal{P}),\mathcal{GT})
\end{equation}
where $\mathcal{GT}$ are the ground-truth labels corresponding to the predicted interaction set $\mathcal{P}$. Focal loss is defined via ${Focal}(p)=-(1-p)^\gamma \log (p)$, where $\gamma$ is set as a hyperparameter. The overall loss is constructed as follows:
\begin{equation}
\mathcal{L}=\lambda_{hoi} \mathcal{L}_{hoi} +\lambda_s^m \mathcal{L}_s^m +\lambda_w^m \mathcal{L}_w^m +\lambda_s^a \mathcal{L}_s^a +\lambda_w^a \mathcal{L}_w^a
\end{equation}

\section{Experiments}

\subsection{Datasets and Evaluation Metrics}
We conducted training and evaluation on the widely used V-COCO \cite{gupta2015visual} and HICO-DET \cite{chao2018learning}, following the established protocols in previous works \cite{li2019transferable, zhang2022efficient}. Due to the limited space, a detailed description of the datasets and evaluation metrics can be found in Supplementary Material.



\begin{figure*}[t]
  \centering
  \includegraphics[width=0.99\linewidth]{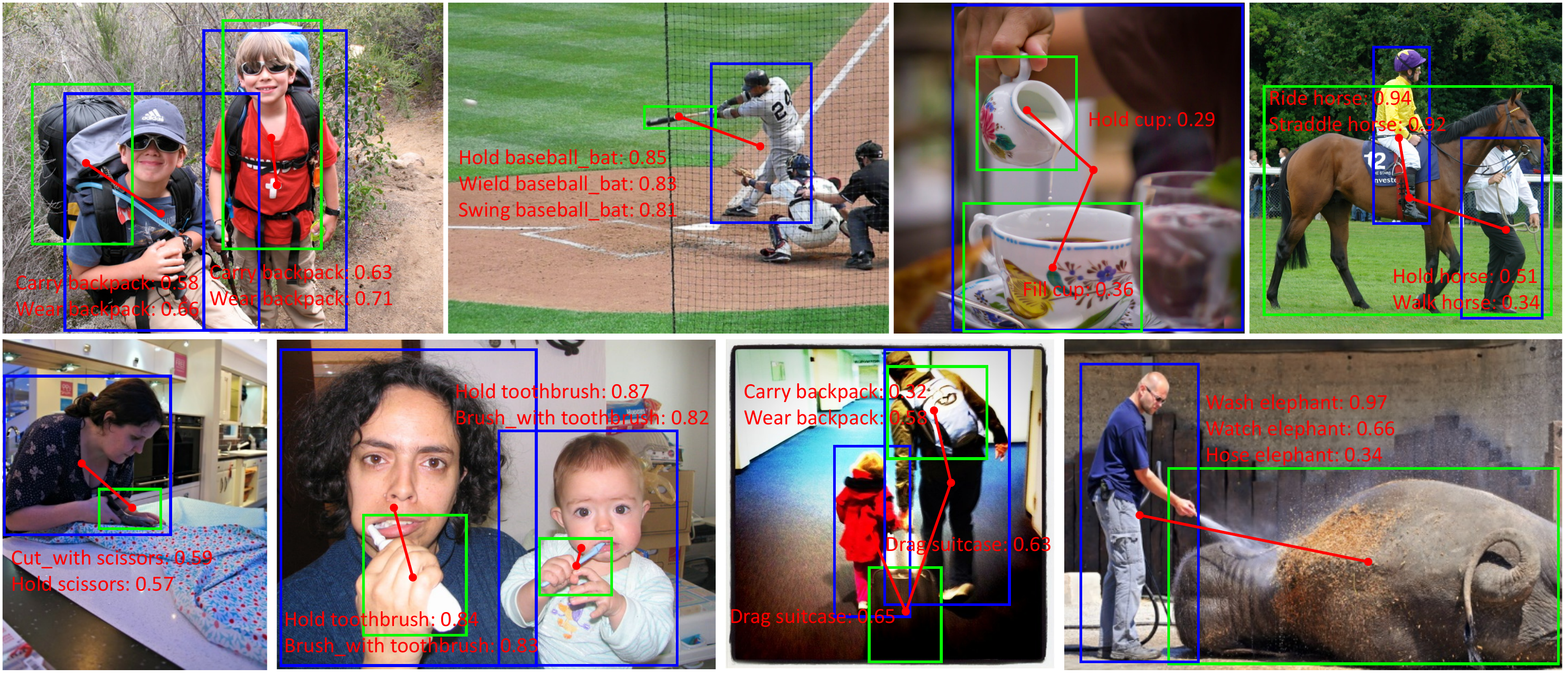}
   \caption{Some representative results of our RmLR method on HICO-DET \cite{chao2018learning} test set.}
   \label{fig:res}
\end{figure*}

\subsection{Implementation Details}
Following the two-stage HOI detector training paradigm \cite{zhang2022efficient}, we first pre-train the DETR on a large-scale image dataset and then fine-tune it on the HICO-DET and V-COCO datasets. For HICO-DET, we initialize the network with DETR pre-trained on MS COCO \cite{lin2014microsoft}. We adopt the data augmentation and preprocessing techniques as \cite{zhang2022efficient}. For cross-modal learning, the number of self-attention and cross-attention layers is set to 2 and 1, respectively. And the dimension of the hidden state in these two mechanisms is set to 1024. For the Focal loss, we set  $\gamma = 0.2$ and $\beta = 0.5$ following \cite{zhang2022efficient}. We also provide a detailed description of implementation details in Supplementary Material.

\subsection{Main Results}
We conducted a comprehensive evaluation of our proposed method in comparison with state-of-the-art HOI methods, such as UPT \cite{zhang2022efficient}, GEN-VLKT \cite{liao2022gen}, and CDN \cite{zhang2021mining}, on the HICO-DET and V-COCO datasets. The results of this comparison are presented in Table \ref{tab2}. Our approach significantly outperforms all previous state-of-the-art methods, and this advantage is maintained across both ResNet-50 and ResNet-101 feature extractors. We also compared our proposed method with some previous methods, such as those relying on extra datasets such as Human Pose \cite{liu2020amplifying} and Vision-and-Language \cite{yuan2022rlip}), by training on larger and richer datasets, as shown in Tables \ref{tab3} and \ref{tab4}. These results demonstrate the superiority of our RmLR method.


\begin{table}[]
\caption{Ablations of different modules of our RmLR Framework on V-COCO [\textcolor{green}{16}]. ``SA'' and ``WA'' indicate sentence- and word-level alignment, respectively. ``KT'' indicates knowledge transfer.}
\centering
\footnotesize
\begin{tabular}{cccccccc}
\hline
\multirow{2}{*}{Variants} & \multirow{2}{*}{IRE} & \multicolumn{2}{c}{IRE-KT} & \multicolumn{2}{c}{IRM-KT} & \multicolumn{2}{c}{V-COCO} \\ \cline{3-8}
         &                    & SA   & WA  & SA   & WA  &${AP}_{role}^{\# 1}$&${AP}_{role}^{\# 2}$ \\ \hline
Plain model     &                     &                  &             &                  &             &58.51    & 63.87\\
w/o CL          &\checkmark           &                  &             &                  &             &61.13    & 67.48\\
w/o Rm          &                     &        &   &\checkmark        &\checkmark   &62.89    & 68.91\\
w/o WA         &\checkmark           &\checkmark        &             &\checkmark        &             &62.37    & 68.29\\
w/o SA         &\checkmark           &                  &\checkmark   &                  &\checkmark   &63.33    & 69.41\\
w/o IRM-KT      &\checkmark           &\checkmark        &\checkmark   &                  &             &62.53    & 68.61\\
w/o IRE-KT      &\checkmark           &                  &             &\checkmark        &\checkmark   &63.42    & 69.49\\
\rowcolor[HTML]{E0E0E0}
RmLR            &\checkmark           &\checkmark        &\checkmark   &\checkmark        &\checkmark   &63.78    & 69.81\\
\hline
\end{tabular}
\label{tab1}
\end{table}

\begin{table}[t]
\centering
\footnotesize
\caption{Experimental results of different Text Encoders. The ResNet-50 \cite{he2016deep} backbone is adopted as the visual feature extractor.}
\begin{tabular}{ccc}
\hline
\multirow{2}{*}{Text Encoder} & \multicolumn{2}{c}{V-COCO} \\ \cline{2-3}
          &${AP}_{role}^{\# 1}$ & ${AP}_{role}^{\# 2}$\\ \hline
ALBERT-base-v2 \cite{lan2019albert} & 63.45  & 69.64       \\
RoBERTa \cite{liu2019roberta}  & 63.49        & 69.62       \\
MobileBERT \cite{sun2020mobilebert}                   & 63.78        & 69.81       \\
BERT-base \cite{devlin2018bert}  & 63.89        & 69.98      \\
\rowcolor[HTML]{E0E0E0}
BERT-large \cite{devlin2018bert}  & 63.93        & 70.05      \\
\hline
\end{tabular}
\label{tab-text-encoder}
\end{table}

\begin{table}[]
\centering
\footnotesize
\caption{FLOPs and Params analysis for HOI detectors on V-COCO \cite{gupta2015visual} dataset with $800\times800$ resolution.}
\begin{tabular}{ccccc}
\hline
Method                                          & Backbone   & MACs (G) & Parms (M) & FPS \\ \hline
\multirow{2}{*}{DETR \cite{carion2020end}}      & ResNet-50  & 57.02    & 36.59      & 29.1  \\
                                                & ResNet-101 & 104.37   & 55.53      & 21.3  \\ \hline
\multirow{2}{*}{UPT \cite{zhang2022efficient}}  & ResNet-50  & 57.11    & 36.86      & 27.5  \\
                                                & ResNet-101 & 104.46   & 55.80      & 20.2  \\ \hline
\multirow{2}{*}{RmLR (Ours)}           & ResNet-50  & 57.22    & 36.98      & 27.2  \\
                                                & ResNet-101 & 105.57   & 55.92      & 19.9  \\ \hline
\end{tabular}
\label{tab-flops}
\end{table}

\begin{table*}[]
\caption{Comparison results with the methods using extra datasets on HICO-DET [\textcolor{green}{6}]. For extra datasets, ``P'' indicates human pose and ``L'' indicates linguistic knowledge.}
\centering
\footnotesize
\begin{tabular}{ccccccccc}
\hline
            &            &        & \multicolumn{6}{c}{HICO-DET}                                                        \\ \cline{4-9}
            &            &        & \multicolumn{3}{c}{Default   Setting} & \multicolumn{3}{c}{Known   Objects Setting} \\ \cline{4-9}
Method (Year)      & Backbone   & Extras & Full       & Rare      & Non-rare     & Full         & Rare        & Non-rare       \\ \hline
PMFNet (2019) \cite{wan2019pose}               & ResNet-50  & L & 17.46      & 15.65     & 18.00  & 20.34  & 17.47  & 21.20  \\
TIN (2019) \cite{li2019transferable}             & ResNet-50  & P & 17.22  &13.51  &18.32  &19.38   & 15.38  & 20.57\\
Peyre et al. (2019) \cite{peyre2019detecting}    & ResNet-50  & P & 19.40  & 14.63     & 20.87   & -  & -  & - \\
FCMNet (2020) \cite{liu2020amplifying}           & ResNet-50  & P+L  & 20.41      & 17.34     & 21.56    & 22.04   & 18.97   & 23.12 \\
PD-Net (2021) \cite{zhong2021polysemy}           & ResNet-50-FPN & L  & 20.76  & 15.68  & 22.28   & 25.59 & 19.93   & 27.28   \\
ACP (2020) \cite{kim2020detecting}               & ResNet-152 & P+L  & 20.59  & 15.92  & 21.98   & - & -   & -   \\
DRG (2020) \cite{gao2020drg}                     & ResNet-50-FPN & P      & 24.53 & 19.47  & 26.04 & 27.98  & 23.11  & 29.43  \\
RLIP-ParSeD  (2022) \cite{yuan2022rlip}          & ResNet-50 & L      & 30.70& 24.67     & 32.50 & -  & -  & -  \\
RLIP-ParSe  (2022) \cite{yuan2022rlip}           & ResNet-50 & L      & 32.84& 26.85     & 34.63  & -  & -  & -  \\
PhraseHOI  (2022) \cite{li2022improving}         & ResNet-50 & L      & 29.29     & 22.03     & 31.46 & 31.97  & 23.99  & 34.36  \\
PhraseHOI  (2022) \cite{li2022improving}         & ResNet-101& L      & 30.03     & 23.48     & 31.99  &33.74  & 27.35  & 35.64  \\
OCN  (2022) \cite{yuan2022detecting}             & ResNet-50 & L      & 30.91     & 25.56     & 32.51  & -  & -  & -   \\
OCN  (2022) \cite{yuan2022detecting}             & ResNet-101& L      & 31.43     & 25.80     & 33.11  & -  & -  & -  \\

\hline
\rowcolor[HTML]{E0E0E0}
RmLR (Ours) & ResNet-50  & L      & 36.93 & 29.03 & 39.29 & 38.29 & 31.41 & 40.34          \\
\rowcolor[HTML]{E0E0E0}
RmLR (Ours) & ResNet-101 & L      & 37.41 & 28.81 & 39.97 & 38.69 & 31.27 & 40.91          \\
\hline
\end{tabular}
\label{tab3}
\end{table*}

\begin{table}[]
\caption{Comparison results with the methods using extra datasets on V-COCO [\textcolor{green}{16}]. }
\centering
\footnotesize
\begin{tabular}{ccc|cc}
\hline

Method & Backbone   & Extras         &${AP}_{role}^{\# 1}$ & ${AP}_{role}^{\# 2}$   \\ \hline
TIN \cite{li2019transferable} &ResNet-50 &P  &48.7&-\\
DRG \cite{gao2020drg} &ResNet-50-FPN &L  &51.0&-\\
FCMNet \cite{liu2020amplifying} &ResNet-50 &P  &53.1&-\\
ConsNet \cite{liu2020consnet} &ResNet-50-FPN &P  &53.2&-\\
RLIP-ParSeD \cite{yuan2022rlip} &ResNet-50 &L  &61.7&63.8\\
RLIP-ParSe  \cite{yuan2022rlip} &ResNet-50 &L  &61.9&64.2\\

\hline
\rowcolor[HTML]{E0E0E0}
RmLR (Ours)                                 & ResNet-50     & L  & 63.78 & 69.81        \\
\rowcolor[HTML]{E0E0E0}
RmLR (Ours)                                 & ResNet-101    & L  & 64.17  & 70.23        \\ \hline
\end{tabular}
\label{tab4}
\end{table}

\subsection{Ablation Studies}
\label{sec:sec4.4}
To illustrate the eﬀectiveness of our proposed approach, we perform ablation studies on each component. Specifically, cross-modal learning contains sentence-level and word-level embedding knowledge distillation for IRE and IRM. The experiments are conducted on the V-COCO \cite{gupta2015visual} dataset with ResNet50 \cite{he2016deep} as the CNN backbone, and the results are reported in Table \ref{tab1}. We also provide an analysis of the computational cost of our method in Table \ref{tab-flops}. The results demonstrate that RmLR achieves a substantial performance improvement while adding only a minor computational cost.

\textbf{The impact of Interactive Relation Encoder.} In ``Plain model'', we follow the typical two-stage HOI detector \cite{zhang2022efficient} to construct a plain model, which directly adopts the entity token features as visual representations and feds them to HOI classifier. For ``w/o CL'', we add IRE for the plain model, but not cross-modal learning. In ``w/o Rm'', we remove the re-mining operation (\emph{i.e.}, IRE) in RmLR to analyze the effect of IRE for the RmLR framework. Since the lack of IRE, we only perform knowledge transfer for IRM in this variant. As shown in Table \ref{tab1}, the introduction of IRE greatly improves the plain model by around 3.1 mAP. And the IRE also shows improvement on RmLR frameworks that are equipped with cross-modal learning.

\textbf{Effect of sentence- and word-level alignment.} For ``w/o WA'' and ``w/o SA'', we remove the word- and sentence-level alignment in the cross-modal learning process. In these two variants, IRE and other settings remained the same. Compared to the complete RmLR, these two variants drop in mAP by 1.5 and 0.4 points, respectively. Adding the word- and sentence-level alignment to ``w/o CL'' variant jointly improves by around 2.5 mAP. Furthermore, the experimental results show that the word-level alignment strategy has a stronger facilitation to cross-modal HOI learning than sentence-level alignment. The possible cause for this phenomenon is that HOI is essentially a variable-size interaction set prediction problem, and a more flexible alignment strategy is beneficial for linguistic knowledge transfer.

\textbf{The impact of transfer position.} In addition, we also verify the necessity of knowledge transfer for IRE and IRM. For ``w/o IRM-KT'' and ``w/o IRE-KT'', we remove the linguistic knowledge transfer for IRM and IRE, respectively. The experimental results show that the performance of these two variants decreased by about 1.2 and 0.3 mAP compared to RmLR. These findings suggest that knowledge transfer for IRM in this architecture is a more efficient approach. Moreover, the results also suggest that distillation for IRE can further improve the performance. Therefore, we chose to perform knowledge transfer for both modules simultaneously, with knowledge distillation for IRM as the primary and IRE as the secondary.

\textbf{Effect of different Text Encoder.}
We build RmLR variants equipped with other text encoders and conduct comparison experiments on the V-COCO dataset to explore the effect of different text encoders. In Table \ref{tab-text-encoder}, we show the results of different text encoders. These results indicate that different text encoders impact HOI recognition capability; generally, larger models may perform better. In addition, all these text models promote our RmLR framework to obtain state-of-the-art results on the V-COCO dataset.

\textbf{The impact of hyperparameters for loss terms.} We also present the results for detailed weight settings for loss function to Table \ref{tab-loss-weight}. The subscript $s$ and $w$ indicates sentence- and word-level alignment loss, respectively. These results demonstrate that our model performance is not very sensitive to the weights of different loss terms.

\subsection{Visualization}
As depicted in Figure \ref{fig:res}, one image may contain multiple individuals and objects, which may or may not interact with each other or engage in several interactions. Hence, we finely aligned and transferred knowledge between visual features and annotation texts and have effectively guided the complex HOI learning process via linguistic prior knowledge. The detection results substantiate the validity of cross-modal alignment and the efficacy of our RmLR approach.

\begin{table}[]
\caption{Experiments on the V-COCO [\textcolor{green}{16}] set \emph{w.r.t} different loss terms. $s$ and $w$ indicates sentence- and word-level alignment loss.}
\centering
\footnotesize
\begin{tabular}{ccccc|cc}
\hline

$\lambda_{hoi}$  &$\lambda_s^m$  &$\lambda_w^m$  & $\lambda_s^a$  & $\lambda_w^a$  &${AP}_{role}^{\# 1}$ & ${AP}_{role}^{\# 2}$   \\ \hline
1.0 &1.0 &1.0  &1.0 &1.0  &62.98&69.11\\
1.0 &1.0 &0.5  &1.0 &0.5  &62.73&68.95\\
1.0 &0.5 &1.0  &0.5 &1.0  &63.35&69.52\\

2.0 &0.5 &0.5  &0.1 &0.1  &63.05&69.14\\
2.0 &1.0 &1.0  &0.08 &0.08 &63.59&69.55\\

2.0 &2.0 &2.0 &0.1 &0.1  &63.57&69.62\\
2.0 &1.0 &1.0 &0.3 &0.3  &63.55&69.69\\

2.0 &2.0 &1.0 &0.5 &0.1  &63.43&69.39\\
2.0 &1.0 &2.0 &0.1 &0.3  &63.69&69.77\\
\rowcolor[HTML]{E0E0E0}
2.0 &1.0 &1.0 &0.1 &0.1  &63.78&69.81\\
\hline
\end{tabular}
\label{tab-loss-weight}
\end{table}

\section{Conclusion}
In this paper, we introduce a systematic and unified framework called RmLR, which leverages structured text knowledge to enhance HOI detector. To address the issue of interaction information loss in the two-stage HOI detector, we propose a re-mining strategy to generate more comprehensive visual representations. We then develop fine-grained sentence- and word-level alignment and knowledge transfer methods to effectively address the many-to-many matching problem between multiple interactions and multiple texts in HOI-VLM. These strategies alleviate the matching confusion problem caused by simultaneous occurrences of multiple interactions, thus improving the effectiveness of the cross-modal learning process in HOI detection filed. Experimental results on the public datasets demonstrate the eﬀectiveness of our approach, which achieves state-of-the-art performance. We hope the proposed RmLR may serve as an architecture guideline for future research in this area.

\section{Acknowledgements}
This work is supported by the National Natural Science Foundation of China under grant 62271143, and the Big Data Center of Southeast University.


{\small
\bibliographystyle{ieee_fullname}
\bibliography{egbib}
}


\newpage

\appendix
\newpage


The supplementary materials are organized as follows. In Appendix \ref{sec:app-motivations}, we elaborate on the motivations behind the development of the RmLR framework. In Appendix \ref{sec:app-Implementation}, we present a more detailed description of our architecture. In Appendix \ref{sec:app-Datasets}, we outline the datasets and evaluation metrics used in our experiments. In Appendix \ref{sec:app-Training-Inference}, we provide an in-depth explanation of the training and inference procedures. In Appendix \ref{sec:app-cases-loss}, we discuss additional cases that demonstrate the interaction loss phenomenon. In Appendix \ref{sec:app-layer-number}, we examine the effects of varying the number of layers in different modules. In Appendix \ref{sec:app-extras}, we implement the Interaction Relation Encoder using a pre-trained human pose detection model and assess its performance. In Appendix \ref{sec:app-relationship}, we explore the connection between our RmLR method and other CLIP-based approaches. In Appendix \ref{sec:app-Visualization}, we present additional detection results for further analysis.

\begin{figure*}
\centering
\includegraphics[width=0.99\linewidth]{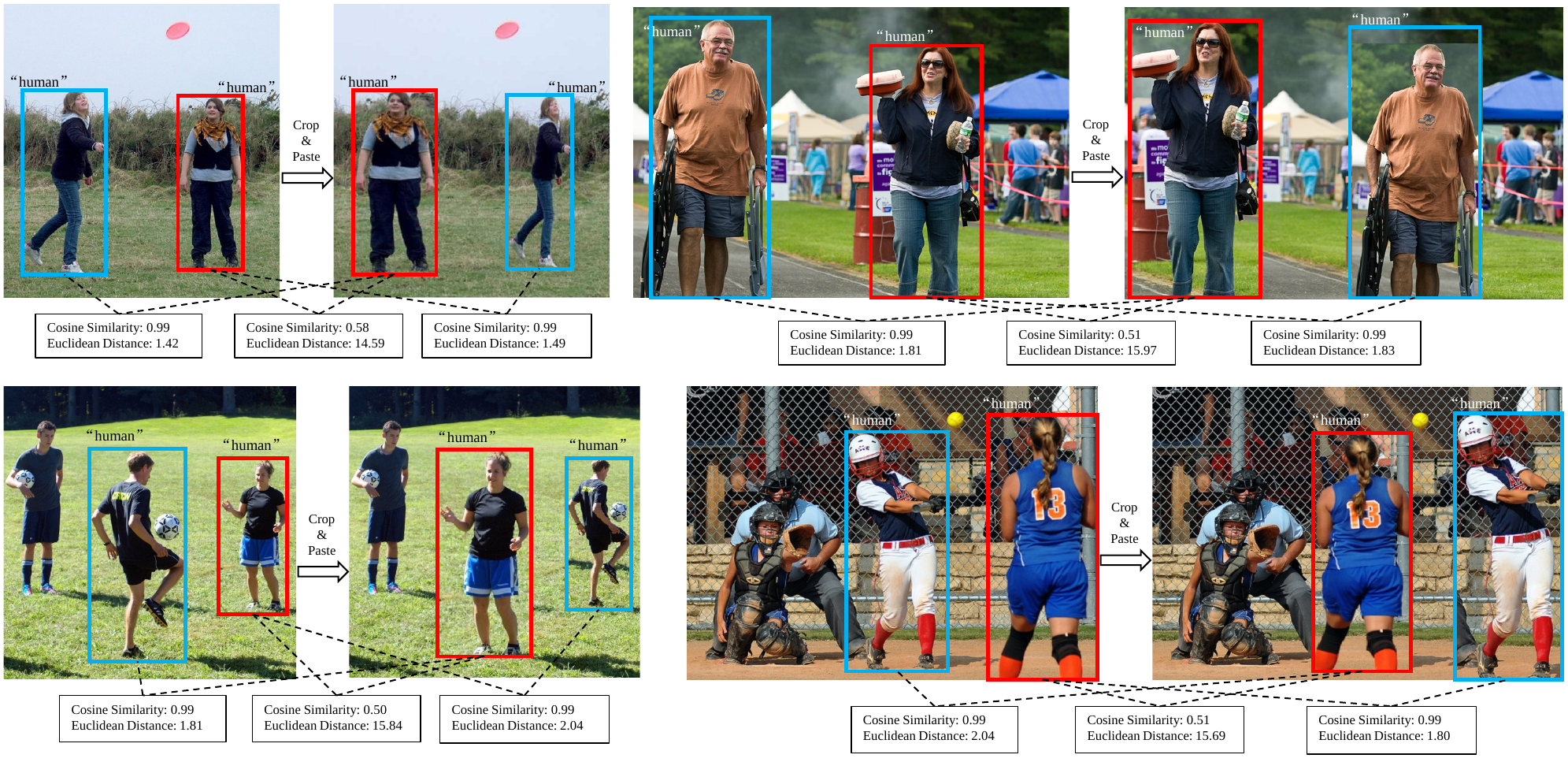}
\caption{We provide further examples to elucidate the phenomenon of interaction information loss in two-stage Transformer-based HOI detectors. Figure \ref{fig10} showcases instances from the HICO-DET and V-COCO datasets, where we evaluate the output tokens of DETR \cite{carion2020end} using both cosine similarity and Euclidean distance metrics. Our results corroborate earlier observations that the output tokens of the detection model predominantly pertain to spatial positioning and object categories, rather than the interaction information. This is exemplified by the fact that individuals situated in the same position exhibit similar features, regardless of the actions they perform.}
\label{fig10}
\end{figure*}

\section{Motivations for Our RmLR Framework}
\label{sec:app-motivations}
An effective HOI detector must concurrently handle both object detection and interaction relation recognition tasks. The latter imposes a more substantial requirement on the model's capability to comprehend visual features. Moreover, optimizing the model by solely mapping the <$person, action, object$> combinations in HOI datasets \cite{gupta2015visual}\cite{chao2018learning} to one-hot labels presents challenges due to the flexibility and diversity inherent in these annotations.

In recent years, several studies have investigated the integration of language prior knowledge from text to guide the learning of HOI models \cite{zhong2021polysemy}\cite{iftekhar2022look}\cite{li2022improving}\cite{wang2022learning}\cite{yuan2022detecting}. Incorporating linguistic modality information has led to modest improvements in the performance of existing HOI methods. However, a majority of these approaches employ a CLIP-like technique to condense the textual semantic features of multiple interaction actions into a fixed-length vector \cite{radford2021learning}. For set prediction problems such as HOI, this compression strategy imposes limitations on the transfer of cross-modal knowledge.

Consequently, we propose a novel cross-modal HOI detection framework that enhances visual feature extraction and cross-modal learning efficiency from two perspectives:

\begin{itemize}
\item Firstly, we perform a qualitative and quantitative analysis of the interaction information loss issue in two-stage visual HOI detectors. We provide supplementary examples in Appendix \ref{sec:app-cases-loss} to corroborate our observations. To tackle this problem, we introduce the Interactive Relation Encoder (IRE), designed to \textbf{re-mine} visual features specifically for HO interaction recognition.

\item Secondly, considering that HOI prediction involves set prediction tasks, we introduce sentence- and word-level alignment strategies to facilitate effective cross-modal \textbf{learning} and ensure knowledge transfer from linguistic modalities.

\end{itemize}
By incorporating these richer multi-modal representations, we can ultimately achieve improved HOI recognition performance.

\section{More Implementation Details}
\label{sec:app-Implementation}
The Visual Feature Extractor and Entity Detection module of our RmLR are based on ResNet \cite{he2016deep} and DETR\cite{carion2020end}, respectively. For the Interactive Relation Encoder and Interaction Reasoning Module, we use 2 and 1 Transformer encoder layers, respectively. We follow the two-stage HOI detector training paradigm\cite{zhang2022efficient}, where we first pre-train DETR on a large-scale image dataset and then fine-tune it on HICO-DET and V-COCO datasets. The weights of DETR remain frozen during fine-tuning. To initialize the network for HICO-DET, we use DETR pre-trained on MS COCO \cite{lin2014microsoft}. However, for V-COCO, we exclude some of COCO’s training images that are contained in the V-COCO test set when pre-training DETR. We use an FC layer to map the global context features to 512-dimensional vectors. Similarly, we use an FC layer to map the output of IRE's interactive feature to the same dimension (512). For the spatial features (entity tokens), we concatenate human and object tokens to construct a 1024-dimensional vector.

We employ the data augmentation and preprocessing techniques proposed in \cite{zhang2022efficient}. Specifically, we resize the input images such that the shorter side is within the range of 480 to 800 pixels and the longer side is limited to 1333 pixels. In our cross-modal learning approach, we use two self-attention layers and one cross-attention layer, with a hidden state dimension of 1024. We set $\gamma = 0.2$ and $\beta = 0.5$ for the Focal loss, following \cite{zhang2022efficient}. To determine new hyper-parameters, we perform cross-validation. We use the Adam optimizer with an initial learning rate of $10^{-4}$ and cosine learning rate decay strategy. Our model is trained with a batch size of 8 for 20 epochs on four 3080 GPUs.

\begin{table}[t]
\caption{Effect of the \#Layers of Different Modules on the V-COCO test set. ``CML-SA'' indicates self-attention layers in cross-modal learning.}
\centering
\begin{tabular}{cccc}
\hline
\multicolumn{2}{c}{\#Layer} & \multicolumn{2}{c}{V-COCO} \\ \hline
IRM  & CML-SA  &${AP}_{role}^{\# 1}$ & ${AP}_{role}^{\# 2}$\\ \hline
1           & 1             & 63.71        & 69.76       \\
1           & 2             & 63.78        & 69.81       \\
2           & 1             & 63.59        & 69.62       \\
2           & 2             & 63.75        & 69.77       \\ \hline
\end{tabular}
\label{tab6}
\end{table}

\section{Details of Datasets and Evaluation Metrics}
\label{sec:app-Datasets}
\textbf{V-COCO} \cite{gupta2015visual}. V-COCO is a popular dataset for benchmarking HOI detection, which is built upon the MS-COCO dataset. The mean average precision (mAP) is used for evaluation. For object occlusion cases, two evaluation scenarios are considered. Scenario 1 (${AP}_{role}^{\# 1}$) considers a strict evaluation criterion that requires the prediction of a null bounding box with coordinates [0, 0, 0, 0], Scenario 2 (${AP}_{role}^{\# 2}$) relaxes this condition for such cases by ignoring the predicted bounding box for evaluation.

\textbf{HICO-DET} \cite{chao2018learning}. We follow the previous methods \cite{ li2019transferable} to evaluate on the HICO-DET. The mAP metric is computed in $Default \, settings$ and $Known \, Objects \, Setting$ for three categories: \textbf{Full} (all 600 HOI classes), \textbf{Rare} (138 classes that have less than 10 training samples), \textbf{Non-rare} (462 classes that have more than 10 training samples). Here the $Default \, setting$ represents that the mAP is calculated over all testing images, while $Known \, Object \, Setting$ measures the AP of each object solely over the images containing that object class.

\section{Details of Training and Inference}
\label{sec:app-Training-Inference}
To guarantee the effectiveness and efficiency of our approach, we systematically design three stages that ensure robust visual feature extraction and successful cross-modal knowledge transfer: (i) \textbf{Re-mining Visual Interaction-Relevant Features}: This stage employs a visual feature extractor and the IRE module to capture low-level features and model interactive relations; (ii) \textbf{Cross-Modal Alignment for Visual and Textual Representations}: This stage devises sentence- and word-level alignment strategies to establish correlations between the semantic information of different modalities; (iii) \textbf{Reasoning Using Linguistic Knowledge}: This stage utilizes an interaction reasoning module to integrate visual and linguistically-enhanced representations.

In this section, we present a comprehensive pseudo-code that outlines the training and inference procedures of RmLR in Algorithm 1. The three stages within this pseudo-code correspond to the three phases previously discussed. For the sake of simplicity, we exclude the training process of the object detection model in the first stage.

\begin{algorithm}[tbp]
    \caption{The training and inference process of RmLR framework.}
    \label{algorithm1}
    \SetAlgoLined
    \SetKwInOut{Input}{Input}
    \KwIn{Pre-trained object detector, pre-trained text encoder \cite{sun2020mobilebert}, maximum training epochs $N$. }
    Init $\tau = 0$; \\
    Initialize and freeze the weights of $\mathcal{F}_{ED}$ with pre-trained object detector weights;\\
    \While{$\tau \leq N $}{
    \textbf{1. Learning Visual Features}\\
        (1) Extract the low-level features $\X^v$ for input $\I$; \\
        (2) Flatten and project the $\X^v$ into $\Z^v$;\\
        (3) Entity Detection: $(\mathcal{S}^v, \mathcal{B}^v, \mathcal{C}^v)=\mathcal{F}_{E D}\left(\Z^v , \mathcal{Q}_o\right)$;\\
        (4) Exhaustively generate HO pairs and filter away invalid combinations;\\
        (5) Obtaining pair-wise token features $\tilde{S}^v$;\\
        (6) Interactive relation modeling via Transformer encoder layer: $\mathcal{X}^v_e=\F_{enc}\left(\X^v\right)$;\\
        (7) Masked RoI operation is adopted to generate union region features $m^v$;\\
        (8) Calculate the global context feature $g^v$;\\
        (9) Concatenate the $[g^v,\tilde{s}^v,m^v]$ to obtain overall visual features for HO pairs;\\

    \textbf{2. Learning Cross-modal Content}\\
        (1) Serialize annotation labels as sentence $\mathcal{T}$;\\
        (2) Tokenize the $\mathcal{T}$ into $\mathcal{Z}^l$ and then map to $\mathcal{X}^l$;\\
        (3) Calculate the $[CLS]$ tokens and word embeddings: $\left(\mathcal{E}_{c l s}, \mathcal{E}^w\right)=\F_{T E}\left(\mathcal{X}^l\right)$;\\
        (4) Self-attention for the $\mathcal{M}^v$ and $\mathcal{O}^v$;\\
        (5) Associate the HO candidates with annotations to obtain the $\mathcal{M}^{v a}$ and $\mathcal{O}^{v a}$;\\
        (6) Cross-alignment for the two modality representations to obtain $\widehat{\mathcal{M}}^{v a}$ and $\widehat{\mathcal{O}}^{v a}$;\\
        (7) Calculate the L1 loss $\mathcal{L}^{m}$ and $L^{a}$ for IRM and IRE, respectively;\\

    \textbf{3. Reasoning Using Knowledge}\\
        (1) Reasoning using linguistic knowledge enhanced visual features and logits;\\
        (2) Calculate the overall loss; \\
        (3) Optimize the learnable weights of RmLR;
    }
    \KwOut{
        The optimized weights of RmLR.
    }
\end{algorithm}

\begin{figure*}[]
  \centering
  \includegraphics[width=0.99\linewidth]{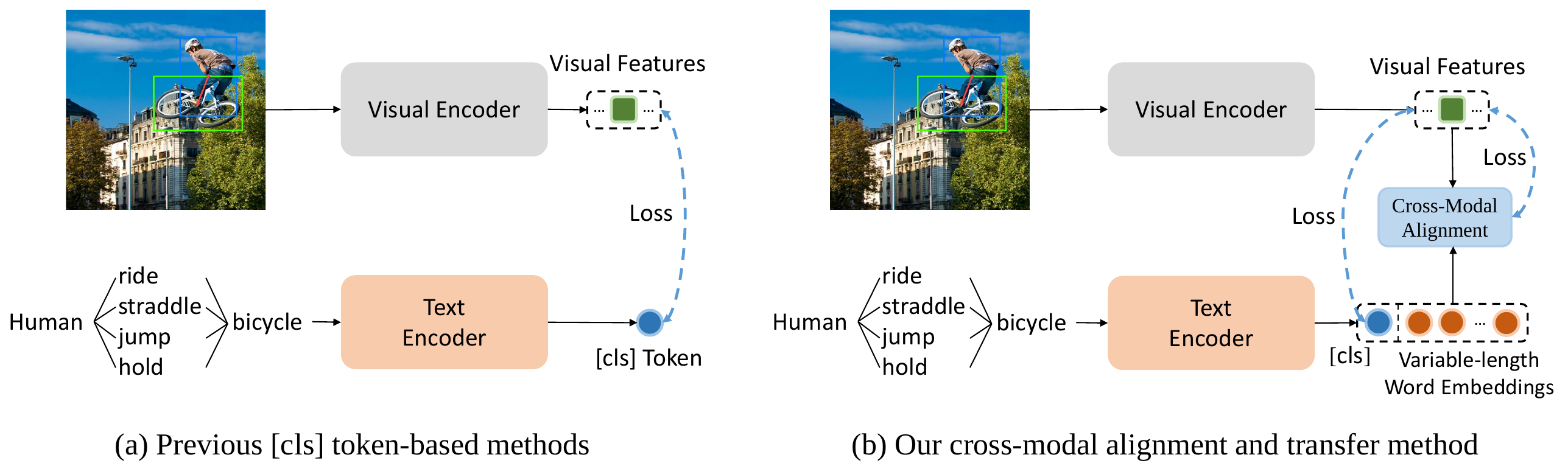}
   \caption{The HOI task involves predicting multiple interaction categories for one human-object pair, making it a set prediction problem. Our RmLR approach employs a more refined knowledge transfer operation compared to the previous HOI-VLM method, which ensures the effectiveness and efficiency of cross-modal learning of HOI detector.}
   \label{fig:4}
\end{figure*}

\begin{figure*}[]
  \centering
  \includegraphics[width=0.99\linewidth]{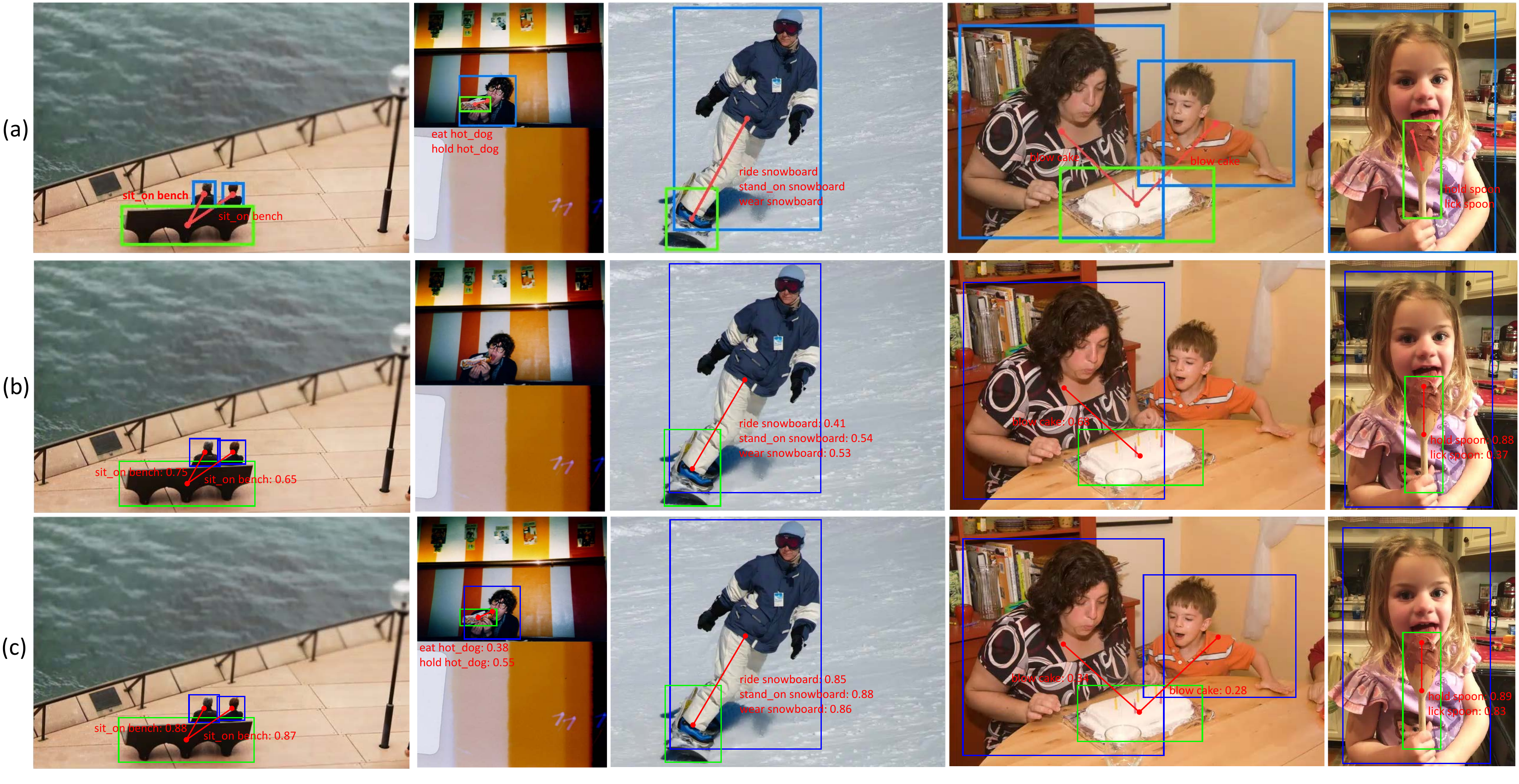}
   \caption{Visualization of HOI annotations and detection results from UPT \cite{zhang2022efficient} and proposed RmLR method. From top to bottom, the images depict the ground truth annotations, UPT results, and our results, respectively. These comparisons reveal that UPT suffers from false negative and low confidence results. In contrast, our RmLR method achieves more accurate and confident HOI detection results.}
   \label{fig:5}
\end{figure*}

\begin{table*}[]
\caption{Performance comparison on the V-COCO test set. The ``Extra Dataset" represent the dataset other than HOI datasets.}
\setlength\tabcolsep{15pt}
\centering
\begin{tabular}{ccccccc}
\hline
           &               & \multicolumn{3}{c}{HICO-DET (Default Setting)} & \multicolumn{2}{c}{V-COCO}\\ \cline{3-7}
Backbone   & Extra Dataset & Full  & Rare  & Non-rare & ${AP}_{role}^{\# 1}$        & ${AP}_{role}^{\# 2}$       \\ \hline
ResNet-50  & -             & 36.93 & 29.02 & 39.29    & 63.78 & 69.81       \\
ResNet-50  & CrowdPose     & 37.15 & 30.18 & 40.23    & 63.93 & 69.97       \\
ResNet-101 & -             & 37.41 & 28.81 & 39.97    & 64.17 & 70.23       \\
ResNet-101 & CrowdPose     & \textbf{38.29} & \textbf{31.05} & \textbf{40.37}    & \textbf{64.38} & \textbf{70.45}       \\
\hline
\end{tabular}
\label{tab8}
\end{table*}

\section{More Cases about Interaction Information Loss Phenomenon}
\label{sec:app-cases-loss}
In the main paper, we have proposed that two-stage Transformer-based HOI detectors tend to lose interactive information. In this section, we present additional evidence to support this claim. Figure \ref{fig10} shows more examples, where we measure the output tokens of DETR \cite{carion2020end} not only with cosine similarity but also with Euclidean distance. The results obtained using Euclidean distance also support the conclusion drawn in Figure \ref{fig1}, that the output tokens of the detection model are only related to position information. These results further reinforce the claim that the two-stage HOI detectors suffer from a loss of interactive information.

\section{Selection of \#Layers of Different Modules}
\label{sec:app-layer-number}
In this section, we present a comprehensive comparison of the number of layers among various models. To fully unleash the potential of our method, we also conducted experiments to compare the performance of different numbers of layers in the IRM module, as presented in Table \ref{tab6}. Moreover, we also investigated the effect of varying the number of self-attention layers in cross-modal learning. Our results demonstrate that the performance improvement of the model is constrained by only increasing the number of layers in IRM and CML-SA.

\section{Modified IRE module using Human Pose Information}
\label{sec:app-extras}
In Section \ref{sec:sec3.2} and Table \ref{tab1}, we presented the need for re-mining interaction-relevant information in two-stage HOI detectors. In our proposed RmLR framework, the IRE module is a learnable component for interactive relationship modeling, gradually acquiring the ability to capture HO interaction cues under HOI annotation and textual semantic information supervision. Furthermore, we replace the IRE module with an explicit human posture recognition model to learn the union interaction feature of HO candidates. This model is pre-trained on the CrowPose \cite{li2019crowdpose} dataset, and we freeze its weight for model training and reasoning as an explicit interaction learning module. We conducted comparative experiments on ResNet-50 and ResNet-101-based RmLR on two datasets, and the results in Table \ref{tab8} show that the IRE module trained with additional datasets can further enhance the RmLR framework. This finding also confirms the necessity of re-mining interaction features from a different perspective.

\section{Relationship between RmLR with other HOI-VLM Methods}
\label{sec:app-relationship}
As discussed in Section \ref{sec:sec2.3}, current state-of-the-art HOI-VLM methods can be categorized into two groups: VLP-based and knowledge distillation-based approaches. VLP methods typically rely on large-scale Vision-and-Language datasets for cross-modal pre-training and fusion of text and image features. In contrast, our proposed method falls under the knowledge distillation category, which enhances the optimization of visual models by transferring knowledge from pre-trained language models.

Our approach is innovative in two key aspects compared to existing HOI-VLM methods:

Firstly, most current knowledge distillation-based methods are somewhat simplistic, such as some CLIP-based HOI detection methods \cite{wang2022learning}\cite{liao2022gen}. These methods directly map annotation text to a fixed-length feature vector and use it to guide the visual model in learning semantic information. While they have achieved some success in exploring HOI-VLM, they still suffer from several drawbacks that need to be addressed. The HOI task is essentially a set prediction problem, where an image may contain multiple HO pairs with various interactions within each pair. Our experimental results in Section \ref{sec:sec4.4} and Table \ref{tab1} demonstrate that simply compressing the semantic information of these interactions into a fixed-length sentence representation (\emph{i.e.}, [cls] tokens) limits the effectiveness of HOI recognition. This approach constrains the full utilization and effective transfer of linguistic information. Therefore, implementing cross-modal alignment and association from text to visual modality is essential to ensure the successful transfer of linguistic prior knowledge to the visual model.

Secondly, our method differs from the general VLP approach because a large vision-and-language dataset is not required in the training process. The training of the HOI detector can be completed solely through efficient fine-tuning and knowledge transfer on the HOI dataset. Furthermore, our RmLR method exhibits exceptionally high training efficiency on HOI datasets. Based on a four 3080 GPU server, the training process of the ResNet-50-based RmLR model takes only about 1.5 hours on the V-COCO dataset and about 12 hours on the HICO-DET dataset.

\section{Visualization for the HOI Detection}
\label{sec:app-Visualization}
We present visualizations of HOI annotations and detection results on the HICO-DET \cite{chao2018learning} test set in Figure \ref{fig:5}. The annotations in (a) demonstrate that an image may contain multiple HO pairs, and various interactions may occur within a single HO pair. Therefore, HOI detectors must predict an HO pair and interaction category set. The detection results of the UPT and our RmLR methods are shown in (b) and (c), respectively. These results reveal that the UPT method \cite{zhang2022efficient} is susceptible to false negatives and low confidence results. Even for some obvious interactions, the UPT method produces highly fluctuating prediction confidence. In contrast, our method achieves more accurate results for both HO pair and interaction category set prediction. These visualizations reinforce the quantification results presented in Table \ref{tab2}, suggesting that our RmLR framework possesses a significantly stronger interaction understanding capability.

\end{document}